# aSTDP: A More Biologically Plausible Learning

Shiyuan Li

## Abstract

Spike-timing dependent plasticity in biological neural networks has been proven to be important during biological learning process. On the other hand, artificial neural networks use a different way to learn, such as Back-Propagation or Contrastive Hebbian Learning. In this work we introduce approximate STDP, a new neural networks learning framework more similar to the biological learning process. It uses only STDP rules for supervised and unsupervised learning, every neuron distributed learn patterns and don't need a global loss or other supervised information. We also use a numerical way to approximate the derivatives of each neuron in order to better use SDTP learning and use the derivatives to set a target for neurons to accelerate training and testing process. The framework can make predictions or generate patterns in one model without additional configuration. Finally, we verified our framework on MNIST dataset for classification and generation tasks.

## 1. Introduction

All animals with brains has the ability to learn, but how they learns is still being studied by neuroscientists. Spike-timing dependent plasticity (STDP) [4, 13, 16] is believed to be the fundamentals of brain's learning process. Although it can also be described in very simple mathematical forms [1], but which is very different from common learning algorithms for artificial neural networks (ANN), like Back-Propagation (BP) [2], Contrastive Hebbian Learning (CHL) [5] or Simulated Annealing [18].

There are many differences between biological learning and learning algorithms of ANN. Some properties that biological learning has but ANN learning algorithms don't are: (i) All the information a neuron needs for learning comes from the neuron itself, like input and output of the neuron or derivatives. They don't require other neuron's behavior or a global loss, neither the information to tell them when to learn, like clamped phase or free phase in CHL. (ii) They don't distinguish between supervised learning or unsupervised learning, just use one learning rule for all situations. They also don't distinguish between training and testing. (iii) Some observed biological characteristics, like feedback connections, asymmetric weights and temporal dynamics of neurons.

There are many ANN Learning algorithms that can satisfy some of the properties above. Contrastive Hebbian learning (CHL) [5, 14, 15] does not require knowledge of any other neurons or a global loss, but CHL has to tell the network to do different algorithms in clamped phase and free phase and it also requires synaptic symmetries. Target propagation [12] computes local errors at each layer of the network using information about the target, which not need a global loss but still need the derivatives of other neurons to propagated error signal. The recirculation algorithm [11] don't need any other information but the derivatives of the neurons, unfortunately it need symmetries weight. The framework in [17] use difference as a target to perform back-propagation. It doesn't require global loss and has asymmetries weight, but the learning method is different from real neurons like STDP or Hebbian rules.

In this paper, we proposed a new learning framework called aSTDP, which satisfy all those properties above and it has the same performance as existing ANN learning algorithms. Our method requires only the input of the neuron and derivative of the neuron to perform learning algorithms, which has the same formula as STDP. On the other hand, we can also treat it as a special CHL without free phase. CHL use clamped phase and free phase to approximate the neuron's derivative, but biological neural networks react to input all the time and don't have a free phase. So we use a weakly clamped phase and a strongly clamped phase to approximate the neuron's derivative at each moment to remove the free phase.

We apply our framework on a continuous hopfield neural network (CHNN) for supervised learning and unsupervised learning. As an energy-based model, CHNN also has the problem of getting stuck in local minima. To solve this problem, we add a fake target to each hidden neuron of the network, it performs a random search-like algorithm to help the network find a better state, which makes the network perform better at test time.

Artificial neural networks commonly use different learning algorithms for supervised or unsupervised learning, but we think there is no real supervise learning in biological learning. Animals learn their environment to survive and evolve, but they doesn't have a teacher or any supervise information. Human babies may learn how to talk from their parents, but parents can't teach them how to see and hear. They just learn to adapt to the environment like a unsupervised way. Motivated to create a more biologically plausible learning, we converting supervised learning to unsupervised learning by concatenate the input data and the labels as new data to do unsupervised learning. This automatically make prediction problem an in-painting problem by clamped the input data to get the labels. We can also generate samples from the model by clamped the label to get the input data. And this making input neurons also output neurons, which is consistent with biological observations that photoreceptor neurons also has feedback connections.

The main contributions of this paper is as follows:

1. We propose a new artificial neural networks learning framework that is very similar to biologically learning process, which based on CHNN, and can be used for both supervised and unsupervised learning.

2. We propose a modified STDP learning algorithms and approximate it with a numerical method.

3. By add fake targets to neurons, we help the network getting better results and increasing speed for both training and testing.

Finally, we verified it on MNIST dataset for classification and generation tasks.

## 2. Model and Methods

### 2.1 Model

We use a CHNN in our framework, as a neural network model with feedback connections and lateral connections instead of just feedforward connections, it is more similar to a biological neural network than other models.

### 2.1.1 CHNN

The CHNN we use is same as [6]. Suppose there are several neurons in the network, every neuron has an internal state $s$, every two neurons are connected with a weight, and every neurons has a bias. Classical leaky integrator neural equation is used to calculate neurons behaviors. It follows:

$$s_{i+1} = s_i + \varepsilon(R_i(s) - s_i) \tag{1}$$

where $R_i(s)$ represents the pressure on neuron i from the rest of the network, while $\varepsilon$ is the time constant parameter. Moreover, suppose $R_i(s)$ is of this form:

$$R_i(s) = \rho'(s_i)(\sum_{j \neq i} W_{j,i} \rho(s_j) + b_i) \tag{2}$$

Where $W_{j,i}$ is the weight from the $j_{th}$ neuron to the $i_{th}$ neuron, $b_i$ is the bias of the $i_{th}$ neuron and $\rho$ is a nonlinear function. The purpose of this formula is to go down the energy function, which is also defined by [6]:

$$E(\theta, s) = \frac{1}{2}\sum_i s_i^2 - \frac{1}{2}\sum_{i \neq j} W_{i,j}\rho(s_i)\rho(s_j) - \sum_i b_i \rho(s_i) \tag{3}$$

Where $\theta$ is the parameters in the model, in this case $W$ and $b$. Derive $E$ with respect to $s$ and with (1), we can get:

$$\frac{ds}{dt} = -\varepsilon \frac{\partial E}{\partial s}(\theta, s) \tag{4}$$

So the dynamics of the network is to perform gradient descent in $E$, each fix point for $s$ will correspond to a local minimum in $E$.

### 2.1.2 Add inputs

We split $s$ into two parts $s = (s_{vis}, s_{hid})$, visible neurons $s_{vis}$ and hidden neurons $s_{hid}$. Where visible neurons contain input and output neurons, and others are hidden neurons. For the input and output neurons $s_{vis}$, we add another pressure to push it towards a target $t$ which is set to the input or output data:

$$\frac{ds_i}{dt} = \beta(t_i - s_i) \tag{5}$$

$$t_i = data_i$$

Where $data_i$ is the input or output data for the $i_{th}$ visible neuron. $\beta$ is the parameter of the degree to which the network is affected by the data. This is also equivalent to add another term in the energy function $E$:

$$C(data, s) = \frac{1}{2}\sum_{i \in s_{in}} (t_i - s_i)^2 \tag{6}$$

Which is the same idea as [7], the different between us is that $\beta$ in this paper is not infinite for input neurons, and we don't distinguish input and output.

So the training process is to make $C$ smaller, and we can define:

$$F(\theta, s) = E + \beta C$$

$$= \frac{1}{2}\sum_i S_i^2 - \frac{1}{2}\sum_{i \neq j} W_{i,j}\rho(s_i)\rho(s_j) - \sum_i b_i\rho(s_i) + \frac{1}{2}\beta\sum_{i \in s_{vis}}(data_i - s_i)^2 \qquad (7)$$

## 2.2 STDP Learning

STDP is considered the main form of learning in brain, it relates the change in synaptic weights with the timing difference between spikes in postsynaptic neurons and presynaptic neurons. Experimental in [1] show that the STDP rule can also be form as:

$$\frac{dW_{i,j}}{dt} = \alpha\rho(s_i)\frac{ds_j}{dt} \qquad (8)$$

Where **α** is the learning rate. If we change the form of STDP rule to:

$$\frac{dW_{i,j}}{dt} = \alpha\rho(s_i)\frac{d\rho(s_j)}{dt} \qquad (9)$$

Which is more similar to the CHL rule since:

$$\Delta W_{ij} + \Delta W_{ji} = \rho_c(s_i)\rho_c(s_j) - \rho_f(s_i)\rho_f(s_j)$$

$$= (\rho_f(s_i) + \Delta\rho(s_i))(\rho_f(s_j) + \Delta\rho(s_j)) - \rho_f(s_i)\rho_f(s_j)$$

$$= \rho_f(s_i)\Delta\rho(s_i) + \rho_f(s_j)\Delta\rho(s_j) + \Delta\rho(s_i)\Delta\rho(s_j)$$

$$\approx \rho_f(s_i)\Delta\rho(s_i) + \rho_f(s_j)\Delta\rho(s_j)$$

Where **ρ_c** is the clamped phase fix point and **ρ_f** is the free phase fix point, we ignore the small term **Δρ(s_i)Δρ(s_j)** and we can get (9) for two interconnected neurons.

We can find that STDP and CHL have the same form, but neither is biologically plausible learning. CHL usually randomly initialize the state each time new data is input which biological neural networks don't. And CHL needs free phase, while biological neural networks also don't. STDP on the other hand is more biologically plausible, it don't have any phases and learn at every moment. But STDP will have a zero **ds** when the state of the network is stable, so it can't even learn a dataset with one data in it. A more biologically plausible should always on clamped phase with a **β** and learn at every moment. When new data is input they will move from the old state to the new state by **β** instead of reinitialize the state.

## 2.2 Calculate derivative

From (7) we can have the derivative:

$$\frac{\partial C}{\partial W} = \frac{\partial C}{\partial s}\frac{\partial s}{\partial W} \qquad (10)$$

Unfortunately **s** is not a free parameter so we can't directly calculate it. Also from (7) we can know, when increase **β**, the network will stop in a state with smaller **C** and bigger **E**, so decreasing **C** is equivalent to increasing **β**. And when **W** is fixed, **s** is a function of **β**. So we can use numerical methods approximate the derivative of **∂C/∂β** by:

$$\frac{\partial C}{\partial \beta} = \frac{\partial C}{\partial s}\frac{\partial s}{\partial \beta} \tag{11}$$

Treat $\Delta s$ as an intermediate bridge we can replace $\partial s$ in $\partial s/\partial \beta$ with $\partial s$ in $\partial s/\partial W$, if we define $\Delta s_\beta = \partial s/\partial \beta * \Delta \beta$ and $\Delta s_W = \partial s/\partial W * \Delta W$, we can get $\partial C/\partial W$ by:

$$\frac{\Delta C}{\Delta W} = \frac{\Delta C}{\Delta s_\beta}\frac{\Delta s_W}{\Delta W} \tag{12}$$

Where $\Delta C/\Delta s_\beta$ is calculate with numerical methods use a small $\Delta \beta$, and $\partial s_W/\partial W$ can get by analytical methods:

$$\frac{\partial s_i}{\partial W_{i,j}} = \rho(s_j) \tag{13}$$

$$\frac{\partial s_i}{\partial B_i} = 1$$

So by calculating the $\partial s/\partial \beta$ corresponding to each $s$, we can reduce the $C$ of the network.

### 2.2.1 Approximate derivative

We add another network in order to more efficiently use numerical methods to approximate $\partial s/\partial \beta$. So there are two networks $net_l$ and $net_s$ in our framework, they have the same structure and parameters $W$ and $B$, but different states $s$ and $\beta$. One with a small $\beta_s$ and other with a large $\beta_l$, this will make the network have two different states $s_s$ and $s_l$. So by calculate the difference between $s_s$ and $s_l$, we can estimate $d\beta/ds$ by:

$$\frac{\partial s}{\partial \beta} \approx \frac{s_l - s_b}{\Delta \beta} \tag{14}$$

Let $z = \rho(s)$, then we can get a new STDP rule:

$$\Delta W_{i,j} = \alpha \rho(s_{s,j})\Delta z_i \tag{15}$$

$$\Delta b_i = \alpha \Delta z_i$$

$$\Delta z = \rho(s_l) - \rho(s_s)$$

Since $\beta_s$ and $\beta_l$ are constants, we can merge $\Delta \beta$ to $\alpha$. In training process, we will use (15) to update $W$ and $B$ at every iteration.

CHL will make each data corresponds to a state, ignoring the relationship between the states, but the relationship between data is important in some tasks like video detection or object tracking. So we also update $W$ and $B$ when changing the state of the network, reducing the energy on the path. Therefore, modeling the relationship between states to a certain extent.

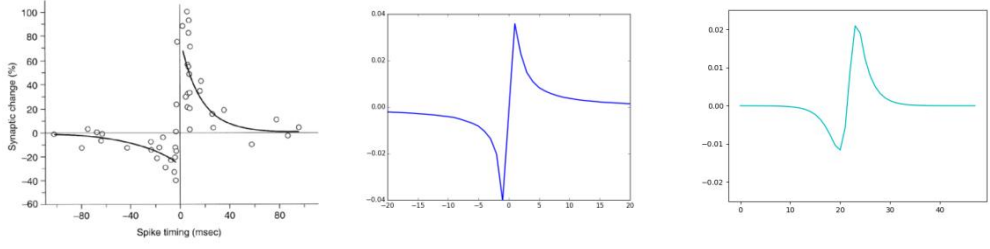

Figure 1 left: biological STDP. Middle: simulation results in [1]. right our simulation result

Our version of STDP rule also can fit the biological observations. Figure 1 left shows the observation of STDP rule in biological nerve cells, and figure 1 middle shows the simulation result by [1] use (8). Notice that the biological STDP is obviously not symmetrical up and down, while the simulation results in [1] are symmetrical up and down. Figure 1 right shows our simulation results, which are also asymmetric up and down since our two *β* has a bias, and more closely with biological observations. The detailed simulation process will be described in section 3.

### 2.2.2 Fake Target

As an energy-based model, CHNN has many stats, each corresponds to a local minima in space of *s*. If the state can interpret the data well, that means the value of the visible neuron is equal to the data. *C* will be zero and *β* will no longer functional, so the $s_s$ and $s_b$ will also be same, which mean *dz* of each neuron will becomes zero. On the other hand if the state can't interpret the data, that means the value of the visible neuron is different to the data. $β_s$ and $β_l$ will cause different $s_s$ and $s_l$ for every neuron in the network and *dz* will be nonzero for each neuron. So, by that *dz* becomes an indicator of network's certainty, finding a good state with small *dz* is critical to the performance of the network.

binocular rivalry [20] is a biologically phenomenon of visual perception in which perception alternates between different images presented to each eye. Biological neural network changes in different states if it can't interpret the data. But as an energy-based model, CHNN will stay in a local minima forever if data don't change.

For a more biologically similar framework, we add a temporary fake target to each hidden neuron to help the network jump out of bad local minima. The value of the fake target is as fallow:

$$t_f = sign(dz) * t_c \qquad (16)$$

Where **sign(dz)** is a function equal to 1 when *dz* is positive and -1 when *dz* is negative, $t_c$ is a constant hyper-parameter. The probability $P_f$ of fake target to acting on neuron i is proportional to the magnitude of *dz*, which fits a Bernoulli distribution, for neuron i:

$$P_{f,i}(t_i = t_{f,i}) = k dz_i \qquad (17)$$

Where *k* is a scale factor, therefore, the larger the *dz*, the more likely the network will jump out of the current local minimum. How to choose a suitable *k* is a tricky problem, if *k* is too small, fake target will be no effect. If *k* is too large, the network will tend to learn the fake target for hidden neurons instead of the real target for visible neurons, and bring instability to training process. To solve these problems, we use a adaptive method to get *k*.

Finding the global minimum in a non-convex space is a hard problem, one method is first use random search to find multiple initial values, then use the gradient-based method to find the local optimum value, and finally compare all the results to find the smallest value. Borrow this idea, we use fake target to jump out of a local minima, and use gradient descent to go to a new local minima, while recording the smallest *dz*. If *dz* in the new local minima is larger than the smallest *dz*, we keep jumping, otherwise we stay.

We keep record smallest *dz* by calculate moving average of *dz*, because network will stay longer in a local minima with a small *dz* than other, it make moving average of *dz* a biased estimate of smallest *dz*. So the probability of adaptive fake target is as:

$$P_f(t = t_f) = \frac{k}{dz_m} dz \tag{18}$$

$$dz_m = m\,dz + (1-m)dz_m$$

Still, *k* is a scale factor and *m* is moving average factor.

We can also think of this process as a kind of random annealing, where *dz* can be thought of as another form of energy. Each neuron randomly reduces its own *dz* to make the overall *C* smaller. Moving average of *dz* can be seen as the temperature.

There are some other benefits by adding fake target:

(1) In order to correctly estimate *ds/dβ*, *dβ* = *β$_l$* - *β$_s$* should be small, but in practice *dβ* cannot be very small. For very deep networks, even a small *dβ* can cause the *s* in the higher layers of *net$_l$* and *net$_s$* to fall into completely different states, this will collapses the training process. By add a fake target, when *dz* is too large, it will force *net$_l$* and *net$_s$* to fall into same states, to fix the training problem.

(2) Due to vanishing gradients, deeper network takes more time to reach a stable state. Fake target will amplify the gradient, which solves the vanishing gradients problem and makes the network reach a stable state faster.

(3) Batch Normalization is a widely used technique for training neural networks, which allowing neurons to have a better dynamic range, make the network to converge faster. But this technique cannot be directly applied to CHNN, because it will make the network unstable. Fake target will also make neurons to have a better dynamic range from *-t$_c$* to *t$_c$*, plays a similar role as Batch Normalization.

### 2.3 Training and testing

For training, we first initialize *W* and *b* randomly, and set *s$_s$* and *s$_b$* to a same randomly selected *s* and as initial. Than we choose a data sample, do a few iteration use (1). We calculate *dz* and update *W* and *b* use (15) at every iteration and also use *dz* to set fake target at every iteration. We loop this process for each data until convergence. The whole algorithms is demonstrate in algorithms 1.

---

Algorithms 1: Approximate Spike-timing Dependent Plasticity, *D* is date set, *ε* is the iteration step for *s* to decrease *E*, *β$_s$* and *β$_l$* is the iteration step for *s* to decrease *C*, *α* is learning rate, *T* is iteration times in each relaxation, *k* is scale factor, *m* is moving average factor.

---

Require: ***D***, ***ε***, ***β$_s$***, ***β$_l$***, ***α***, ***k***, ***m***, ***T***
Initialize ***W***, ***b***, ***s$_s$***, ***s$_l$*** randomly
for n ← 1, . . . , |***D***| do
    ***data*** = ***D$_n$***
    for t ← 1, . . . , ***T*** do
        ***dz*** = ***ρ(s$_l$)*** - ***ρ(s$_s$)***
        ***dz$_m$*** = ***dz*m+dz$_m$*(1-m)***
        Set target for visible neurons by data: ***t$_i$*** = ***data$_i$***, ***β$_{s,i}$=β$_s$***, ***β$_{l,i}$=β$_l$***
        ***u$_i$*** = ***U(0,1)***
        ***I$_{f,i}$*** = ***dz$_i$ > u$_i$*k*dz$_{m,i}$***
        Set fake target for hidden neurons: ***t$_{f,i}$*** = ***sign(dz$_i$)*t$_c$*I$_{f,i}$***, ***β$_{s,i}$=β$_s$*I$_{f,i}$***, ***β$_{l,i}$=β$_l$*I$_{f,i}$***
        Update ***s$_s$*** use (1) and (5): ***s$_s$ = s$_s$ + ε(R$_s$-s$_s$) + εβ$_s$(t-s$_s$)***
        Update ***s$_l$*** use (1) and (5): ***s$_l$ = s$_l$ + ε(R$_l$-s$_l$) + εβ$_l$(t-s$_l$)***
        Update ***W***, ***b*** use (15): ***w = w + α*dz*ρ(s$_s$)***
    end for
end for

---

Notice that we don't reinitialize ***s*** after data changed, so there is not any global signal for neurons. We only change the input of the network, and the rest is left to the network to learn by itself, which is consistent with biological neural networks.

### 2.3.1 Unsupervised learning

Set visible neurons to data, and we can do unsupervised learning. The framework works like an auto encoder, the algorithms is same to algorithms 1.

### 2.3.2 Supervised learning

For supervised learning we split visible neurons into input neurons and output neurons. For each data we first clamp input neurons, hidden neurons and let free output neurons to do a few iterations, we call this free phase. After ***i$_f$*** iterations we clamp input neurons, output neurons and hidden neurons and do another few iterations, we call this champed phase. The algorithm is very similar to algorithms 1, except a little modification.

---

Algorithms 2: Supervised learning version of Approximate Spike-timing Dependent Plasticity, ***D*** is dateset, ***L*** is label dataset, ***ε*** is the iteration step for ***s*** to decrease ***E***, ***β$_s$*** and ***β$_l$*** is the iteration step for ***s*** to decrease ***C***, ***α*** is learning rate, ***T$_f$*** is free phase iteration times in each relaxation, ***T*** is total iteration times in each relaxation, ***k*** is scale factor, ***m*** is moving average factor.

---

Require: ***ε***, ***β$_s$***, ***β$_l$***, ***α***, ***k***, ***m***, ***T$_f$***, ***T***
Initialize ***W***, ***b***, ***s$_s$***, ***s$_l$*** randomly
for n ← 1, . . . , |***D***| do
    ***data*** = ***D$_n$***
    ***label*** = ***L$_n$***
    for t ← 1, . . . , ***T*** do
        ***dz*** = ***ρ(s$_l$)*** - ***ρ(s$_s$)***

>        $dz_m = dz*m+dz_m*(1-m)$
>        Set target for input neurons by data: $t_i = data_i$, $β_{s,i}=β_s$, $β_{l,i}=β_l$
>        if t > $T_f$ then
>            Set target for input neurons by label: $t_i = label_i$, $β_{s,i}=β_s$, $β_{l,i}=β_l$
>        end if
>        $u_i = U(0,1)$
>        $I_{f,i} = dz_i > u_i*k*dz_{m,i}$
>        Set fake target for hidden neurons: $t_{f,i} = sign(dz_i)*t_c*I_{f,i}$, $β_{s,i}=β_s*I_{f,i}$, $β_{l,i}=β_l*I_{f,i}$
>        Update $s_s$ use (1) and (5): $s_s = s_s + ε(R_s-s_s) + εβ_s(t-s_s)$
>        Update $s_l$ use (1) and (5): $s_l = s_l + ε(R_l-s_l) + εβ_l(t-s_l)$
>        Update $W$, $b$ use (15): $w = w + α*dz*ρ(s_s)$
>    end for
> end for

---

In free phase, the network will move to a state $s_f$ with low $C$ of input data, and in clamped phase, the network will move to a state $s_c$ with low $C$ of input data and output data. If $s_f$ is equal to $s_c$, it is the result we want, but if $s_f$ is different from $s_c$, the movement from $s_f$ to $s_c$ will lower the energy of $s_c$ and rise the energy of $s_f$. With enough training, $s_f$ will be replaced by $s_c$.

### 2.3.3 Self-supervised learning

Models such as BERT [21] demonstrate the effectiveness of self-supervised learning in natural language processing. But self-supervised learning in computer visions has not been very successful due to ambiguity in images. BiET [22] uses sparse coding method to convert the image problem into a sequence problem, but the model who used is not a convolutional neural network which is more suitable for processing images.

Since energy-based models are inherently able to resolve ambiguity, we can directly use convolutional neural network-like models for self-supervised learning in our framework. By use part of data as input and rest of data as label, self-supervised learning is same to supervised learning in our framework. The algorithms is same to algorithms 2.

We cover parts of the image and let the network predict the covered parts from the uncovered parts. In free phase, the network may move to a state in one of three situations:

(1) The output is the covered part.
(2) The output is meaningless.
(3) The output is not the covered part, but also some meaningful result.

If in situation 1, this is what we want. If in situation 2, the clamped phase will rise the energy of the state to remove it. If in situation 3, the clamped phase will also rise the energy of the state, but since it is some meaningful state, there must be a data corresponding to the state that can lower it's energy back. So after training, only meaningless state is disappear.

### 2.3.4 Generation

By randomly initializing $s$ and do a few iterations, we can generate data in our framework. By swapping input and output data of some classification task, we can do conditional generation in our framework. The algorithms is same to algorithms 2 except input and output are swapped.

### 2.3.5 Testing

The testing process is same to training process, except the learning rate is zero and could have more iterations.

## 2.4 Other tricks

The are some other tricks use in our framework:

### 2.4.1 Momentum inference

We use momentum to help iterative inference, so we change (1) into:

$$\frac{ds_i}{dt} = \varepsilon v_i \tag{19}$$

Where $v_i$ is the velocity for $s_i$, and we update $v_i$ use:

$$v_{i_t} = m(R_i(s) - s_i) + (1-m)v_{i_{t-1}} \tag{20}$$

Where $m$ is the inertia parameter.

### 2.4.2 Smoothing derivative

We also smoothing the path between different states by make gradients of every point on the path smaller, in order to make the network easier to jump from one state to another state. We perform this by calculate the derivative of $ds$ to $W$, adjust $W$ to make $ds$ smaller.

We have:

$$ds_i = \rho'(s_i)(\sum_{j \neq i} W_{j,i}\rho(s_j) + b_i) - s \tag{21}$$

So the derivative of $ds$ to $W$, $b$ is:

$$\frac{\partial^2 E}{\partial s \partial W_{i,j}} = \rho'(s_i)\rho(s_j) \tag{22}$$

$$\frac{\partial^2 E}{\partial s \partial b_i} = \rho'(s_i)$$

And the learning rule will be:

$$\Delta W_{i,j} = \rho'(s_i)\rho(s_j) \tag{23}$$

$$\Delta b_i = \rho'(s_i)$$

This will make the path between states flatter and easier to jump.

# 3. Experiments

## 3.1 MNIST

In this section, we will verify our algorithm on classification tasks and generation tasks. The dataset we use is the MNIST Handwritten Digit and Letter Classification dataset. We train our model on 60000 numbers for 500000 times and test on 10000 numbers with batch size 128.

The learning rate we use is 0.0001, and we set $β_l$ to 0.4, $β_s$ to 0.25, $ε_l$ to 0.5 and $ε_s$ to 0.4 for

***net*₁** and ***net*ₛ**. Though we can use the same *ε* for both networks, but we use different *ε* to accelerates the training speed. The scale factor ***k*** for fake target is 50 and ***t*_c** of fake target is 0.25. We use sigmoid4 as activate function to speed up the training process, sigmoid4(x) = sigmoid(4x). Parameters ***W*** and ***b*** are initialized with a uniform distributions ***U***(−x, x), x is Xavier initialization parameters $(6/n)^{-2}$, where n is input dims of neuron, and our initialization is not symmetric.

### 3.1.1 Classification

The neural network we use has 784 + 10 visible neurons for images and labels and the number of hidden neurons is 512. Each input and output neuron is connected to hidden neurons, and all hidden neurons are interconnected, but input and output neurons don't connected to each other.

We test unsupervised learning, supervised learning and self-supervised learning on the MNIST dataset. For unsupervised learning, we treat both images and labels as input, and use algorithm 1 for training. In testing time we clamped only input neurons with images and let the network get the label of output neurons, make it works like an in-painting problem. We iterate 160 times for each data in training and testing times. For supervised learning we use algorithm 2 to training the network, and same testing process with unsupervised learning. We iterate 80 times for free phase and 160 times for clamped phase, so there total 240 iterations for each data. For self-supervised learning we first only use images to train the network by cover part of the images by a gray square with 10 pixels in random position, and let network predict the covered part use supervised learning algorithm, the iterations times is same to supervised learning. After that we fix ***W*** and ***b*** for input neurons and hidden neurons, add output neurons and training only output neurons with algorithm 2, the result is showed in table 1.

| Learning method | Network structure | Accuracy |
|---|---|---|
| Unsupervised learning | 784-512-10 | 91% |
| Supervised learning | 784-512-10 | 92% |
| Self-supervised learning | 784-512-10 | 92% |
| Back-propagation MSE | 784-512-10 | 93% |
| Back-propagation cross entropy | 784-512-10 | 97% |

Table 1. Classification result with different learning algorithms

We also use supervised learning to training different network with 1, 2 and 3 hidden layers. Due to the vanishing gradient problem, the inference time of the network will increases exponentially with the depth of the network [7]. But with Fake target the inference time of the network only increases linear with the depth of the network. Table 2 shows the result of different depth of the network, the inference time records in number of steps from data changed to network reach a stable state, which mean ***ds*** is less than 0.001.

| Network architecture | Inference time | Accuracy |
|---|---|---|
| 784-512-10 | 76 | 96% |
| 784-512-512-10 | 95 | 95% |
| 784-512-512-512-10 | 121 | 93% |
| 784-500-500-500-10 no fake target [7] | 500 | 97% |

Table 2. testing time with different network architecture

We use mean-square error (MSE) loss in our framework instead of classification loss like cross entropy, so the result is not as good as standard back-propagation neural network.

### 3.1.2 Generation

We can generate samples by randomly pick a state and do a few iterations, but this usually get meaningless samples. So we use conditional generation method to generate numbers by first randomly pick *s* and then clamped output neurons with label. The label will constrain the distribution of hidden neurons to get more reasonable result, so the network will jump to a number looks image in that label. Figure 2 shows some samples generated by our model.

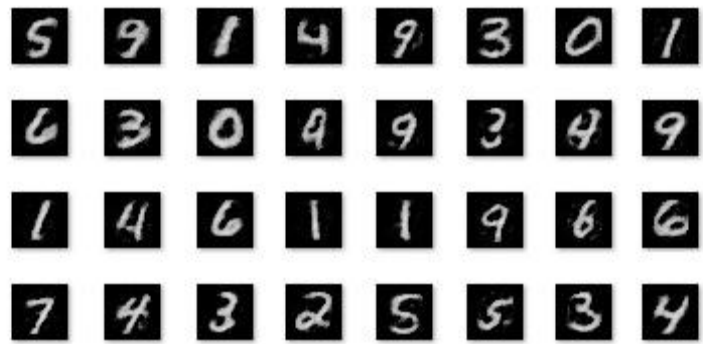

Figure 2 some numbers generated by our model

### 3.2 Toy experiments

We also testing on some features of our framework, like binocular rivalry, neural adaptation, STDP simulations and relationship between states.

### 3.2.2 Binocular rivalry

When biological visual systems see ambiguous input, perception switches between two possible interpretations. For example, when the left eye sees a cat and the right eye sees a dog, it will sometimes perceive the cat and sometimes the dog, toggle back and forth between cat and dog.

To test the binocular rivalry phenomenon in our network, we create a toy dataset with 4 data, the $i_{th}$ data is a four dimensional one-hot vector with $i_{th}$ dimension set to 1 and other set to 0. We train a small network with 4 input neurons and 8 hidden neurons. Then we combine the first data [1, 0, 0, 0] with the second data [0, 1, 0, 0] to make an ambiguous data [1, 1, 0, 0] and feed it into the network. The network starts to oscillate, and the input neurons switch states between two data, as shown in the figure 3.

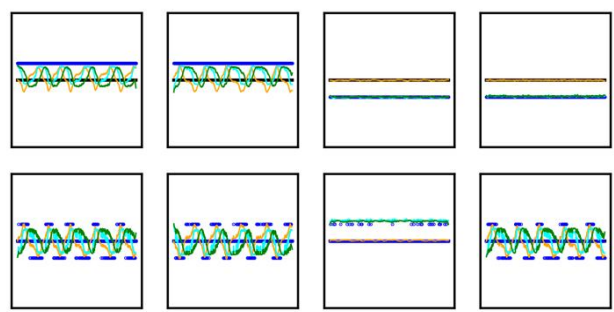

Figure 3 binocular rivalry simulations

Blue dot is the target, green line is $s_s$, blue line is $s_l$ and orange line is $dz$. The first Line of images is input neurons and second line of images is some hidden neurons. We can observe that the main cause of the oscillate is fake target.

### 3.2.2 Neural adaptation

The biological neurons have neural adaptation, when a neuron is activated, it first violently spikes and then drops to a value higher than the inactive state. Also, when a neuron is inactivated, it first strongly suppressed and then increases to a value below the activation state. Faster reflexes are important for wildlife survival, obviously this temporal dynamics can make neurons jump from one state to another state faster. Fake target with momentum inference can lead to the same result. When the input data changes, the network will have a large $dz$, it will trigger fake target and make $s$ get close to the target quickly. Some momentum builds up in the process, when $s$ gets closer to the target, fake target will disappear, but the accumulated momentum will make $s$ keep moving a certain distance in the direction of the target.

Figure 4 shows the similarity between our network and real neurons. Left is the simulation in our paper, right [10] is the response of a real biological neuron to certain features. X axis represents time and Y axis represents intensity, again, blue dot is the target, green line is $s_s$, blue line is $s_l$ and orange line is $dz$.

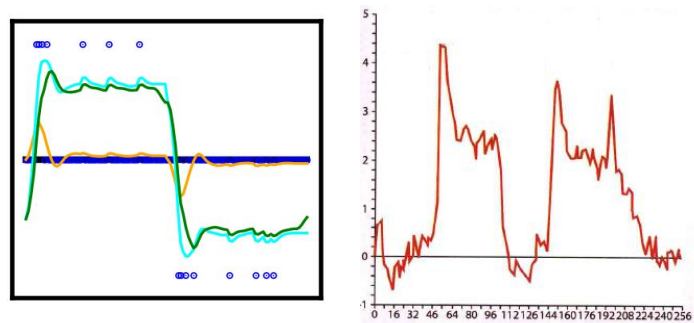

**Figure 4 left: simulations by the model. right: response of a real neurons.**

### 3.2.3 STDP simulations

To simulate STDP in our framework, we built a small network with only two neurons, one input neuron and one output neuron. The **W** and **b** in the network is all zeros to remove the influence between neurons. Biological experiments create presynaptic and postsynaptic potentials through electrodes inserted into neurons, we simulate this behavior by adding a fixed time window target **t=1** to the input and output neuron. Figure 5 shows this process.

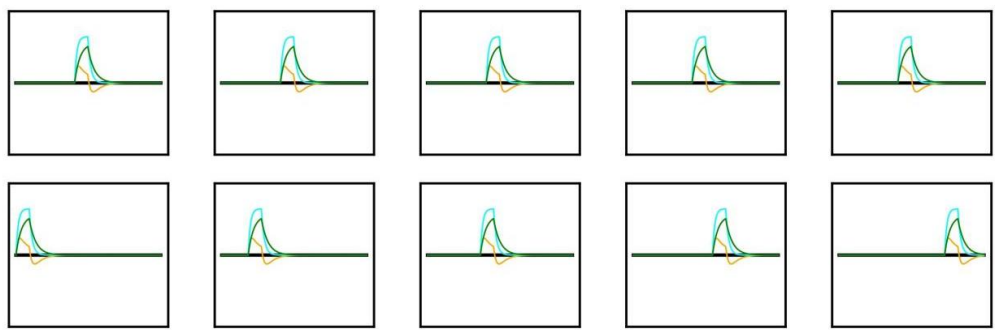

**Figure 5. time relationship between input neurons and output neurons**

First line of images is input neuron, with time window in center. Second line of image is output neuron with same size window but different relative time, from a negative relative time to a positive relative time. We can find that with the appearance and disappearance of the target *t*, the changing of $s_l$ and $s_s$ causes a positive *dz* at first and then negative *dz*. So with (15) a positive input before or after the window will bring a positive or negative change in **W**. With different window size and different *β*, we can get different simulation result in figure 6.

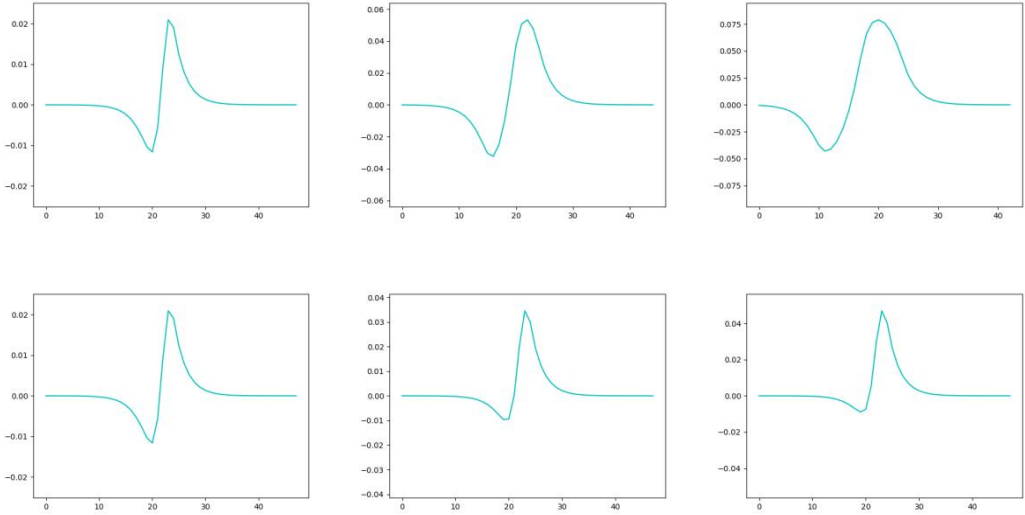

Figure 6. simulation result with different window size and *β*. top left window size 5, top middle window size 10, top right window size 15, bottom left $β_s$ 0.4 and $β_l$ 0.5, bottom middle $β_s$ 0.4 and $β_l$ 0.6, bottom right $β_s$ 0.4 and $β_l$ 0.7

First line of images is same *β* but different window size, second line of images is same window size but different *β*. We can see that as the time window shrinks, the target *t* getting narrow and closer to emitting a single spike, while the curve getting sharper and more close to biological experiments result. And when the proportion of *β* becomes larger, the curve becomes more asymmetric.

### 3.2.4 Relationship between states

To test the ability of the network to model relationships between data, we create a toy dataset with 8 data, the $i_{th}$ input data is a eight dimensional one-hot vector with $i_{th}$ dimension set to 1 and other set to 0, the $i_{th}$ output data is same to $i_{th}$ input data. We train two network on this dataset with same structure, the difference is the order of the data. For the first network we use the adjacent data for the next data, assuming the current data is $i_{th}$ data, the next data is $i_{next}$ = ($i_{th}$±1) mod 8, we call it network81 and data order 81. For the second network we use the adjacent 3 data for the next data, assuming the current data is $i_{th}$ data, the next data is $i_{next}$ = ($i_{th}$±3) mod 8, we call it network83 and data order 83.

| Network | Data order 81 | Data order 83 |
|---|---|---|
| Network81 | 100% | 98% |
| Network83 | 99% | 100% |

Table 3. accuracy of different network and different data order with 160 iterations

We train it with 160 iterations for each data, and test on both network on both data order with input data clamped and let network predict output data. When test on 160 iterations, due to the existence of fake target, the accuracy of the two networks is similar, see Table 1. But when we test on 80 iterations, both networks have better results in their own data order. Obviously, training for a specific order can make the network reach the state faster in a certain data order.

| Network   | Data order 81 | Data order 83 |
|-----------|---------------|---------------|
| Network81 | 100%          | 91%           |
| Network83 | 87%           | 100%          |

**Table 4. accuracy of different network and different data order with 80 iterations**

# 4. Discussion

In this work, we introduced a framework for supervised learning and unsupervised learning, the way it learns is very similar to biological learning. But there are still some possible doubts about our framework.

### 4.1 Why this work

This paper does not prove the convergence of our learning algorithm, but the essence of the learning algorithm in this paper is actually CHL, and the convergence of CHL is been proved. For CHL, the training is mainly divided into two parts, find a fixed point with relatively small *C* for the current data, then make the *C* of this fixed point smaller. This process will continuously bring the fixed point of the network closer to the data, and our learning process is actually equivalent to executing a CHL in each state of the network.

For example, suppose that we execute CHL N times for each data. When the current fixed point does not correspond to the current data, by clamped the input neurons to data with some *β*, CHL will make the network jump to a corresponding fixed point by one CHL clamped phase and free phase. Then when the current fixed point is correspond to the current data, then the standard CHL is executed for remain N - 1 times, it will make the free fix point has lower *C.* When the jump is from one data's fix point into another (for example from $data_i$'s fix point to $data_j$'s fix point), the opposite jump will cancellation the learning (from the $data_j$'s fix point to $data_i$'s fix point). So eventually we make every data's fix point has smaller *C*.

Therefore, after training convergence, the network itself will have many fix points make sense to data with or without input, like humans can imagine things with their eyes closed, which is a property not find in other models [5, 11, 12, 14, 15].

### 4.2 About symmetric weights

We initialize the network with asymmetric weights, and no additional constraints are added during training to make the weights symmetric. Although it has been proved that random feedback can play the same role as back-propagation [8,9], symmetric weights in our framework is important for network to correct propagation error and for fake target to work. However, our framework seems to use a different approach to solve this problem. Observing the network after training, we can find that the weights have changed from asymmetrical to some degree of symmetry. Possible explanation is because positive *s* tend to have positive *dz* and negative *s* tend to have negative *dz*, so the weights of neurons before and after the synapse will be strengthened or weakened at the same time.

On the other hand, the math in (4) has some approximations, the derivative of *E* is actually:

$$\frac{dE}{ds_i} = \rho'(s_i)(\sum_{j \neq i} \frac{1}{2}(W_{j,i} + W_{i,j})\rho(s_j) + b_i) \tag{22}$$

We use $W_{ij}$ instead of $(W_{ij}+W_{ji})/2$ to simplify the calculation.

If we consider $s$ as a force, the space of $s$ will become a force field. The curl of the field is always zero when network has symmetric weights, it causes the particles in the force field to stop. The curl of the field is not always zero when asymmetric weights, particles in the force field may move forever. In each iteration, the learning rule will make the output of the neurons more like the next iteration's output. Because the data is static, the network will eventually tend to be static, which also causes the network to have symmetrical weights.

### 4.3 Corresponding biological explanation

There are still some differences between the framework and real neurons: (i) real neurons has sparse representations. (ii) real neurons respond decreasing to invariant features over time. (iii) real neural networks don't have another network for different $β$.

For (i), we can achieve a similar result by adding a sparse regularization term, but it is unclear whether it is consistent with biological principles. For (ii), biological neural networks have complex chemical processes, our framework is just a simple mathematical simulation, which is somewhat different from reality. For (iii), we think that biological neural networks may have other ways to estimate $dz$, maybe rely on LTP and STP, but the specific process needs future research.